\pdfoutput=1

\documentclass[11pt]{article}

\usepackage{EMNLP2022}
\usepackage[utf8]{inputenc}
\usepackage[T1]{fontenc}    
\usepackage{hyperref}       
\usepackage{url}            
\usepackage{booktabs}       
\usepackage{amsfonts}       
\usepackage{nicefrac}       
\usepackage{microtype}      
\usepackage{amsmath}
\usepackage{natbib}
\usepackage{multicol}
\usepackage{times}
\usepackage{latexsym}
\usepackage{microtype}
\usepackage{inconsolata}
\usepackage{graphicx}
\usepackage{caption}
\usepackage{subcaption}

\title{RL with KL penalties is better viewed as Bayesian inference}
\author{%
  Tomasz Korbak \\
  University of Sussex\\New York University \\
  \texttt{tomasz.korbak@gmail.com} \\
    \And
   Ethan Perez \\
   New York University \\
   \texttt{perez@nyu.edu} \\
   \And
   Christopher L Buckley \\
  University of Sussex\\
   \texttt{c.l.buckley@sussex.ac.uk} \\
}

\begin{document}

\maketitle

\begin{abstract}
Reinforcement learning (RL) is frequently employed in fine-tuning large language models (LMs) to penalize them for undesirable features of generated sequences, such as offensiveness or harmfulness. In this paper, we analyze challenges associated with treating language models as RL policies and show how avoiding those challenges requires moving beyond the RL paradigm. 

We start by observing that naïve maximisation of the standard RL objective unavoidably leads to distribution collapse: turning the LM into a degenerate distribution. Then, we analyze KL-regularised RL, a widely used recipe for fine-tuning LMs, which additionally constrains the fine-tuned LM to stay close to its original distribution. We show that KL-regularised RL is equivalent to variational inference: approximating a Bayesian posterior which specifies how to update a prior LM to conform with evidence provided by the reward function.

Based on these observations, we sketch a Bayesian perspective that provides a first-principles derivation for KL-regularised RL. The Bayesian perspective also separates the modelling problem (defining a target distribution specifying the desired behaviour of an LM) and the inference problem (approximating that target distribution). Finally, it suggests that RL is not a good formal framework for thinking about fine-tuning LMs.
\end{abstract}


\begin{figure}
    \centering
    \includegraphics[width=\linewidth]{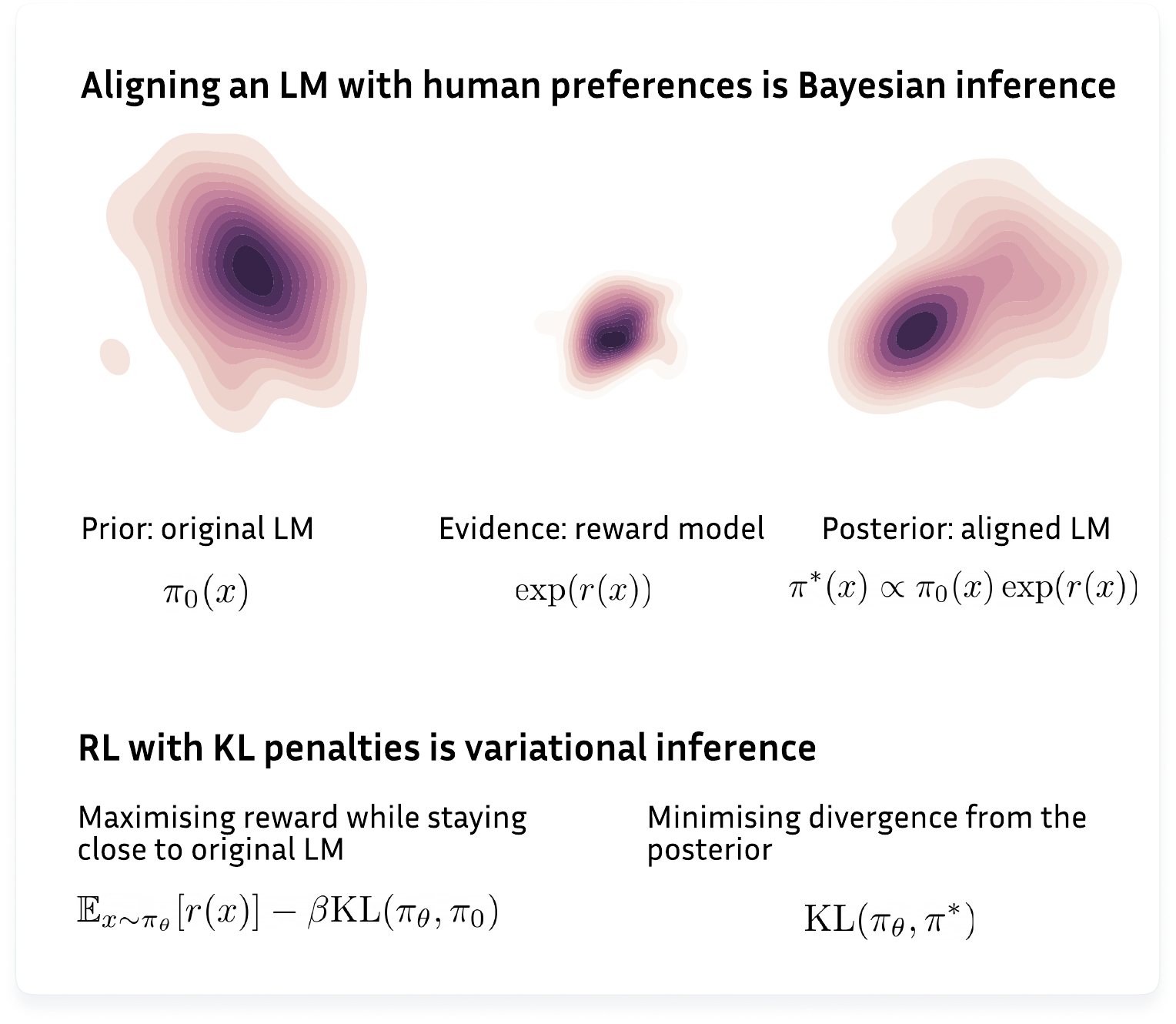} 
    \caption{In the paper, we argue that aligning language models (LMs) with human preferences is a Bayesian inference problem and RL with KL penalties corresponds to solving it via variational inference.}
\end{figure}

\section{Introduction}

Large language models (LMs), such as GPT-3 \cite{gpt3}, tend to generate outputs that reflect undesirable features of their training data such as offensiveness \citep{rtp}, social bias \citep{Bender_parrots}, harmfulness \citep{bai2022training} or dishonesty \citep{lin2021truthfulqa}. Addressing these biases and constraining LMs to be honest, helpful and harmless is an essential part of the problem of aligning LMs with human preferences \citep{lab}. One intuitive approach to aligning LMs is reinforcement learning (RL): capturing human preferences as a reward function and fine-tuning the LM to maximise the reward expected under LM distribution. A practical recipe for implementing this idea is RL from human feedback \citep{Ziegler19}: first, a reward model is trained to predict which of two texts a human prefers and then a pretrained LM is fine-tuned to maximise reward given by the reward model while also being penalized for Kullback-Leibler (KL) divergence from its initial distribution. However, despite immense popularity of RL from human feedback \citep{10.5555/3495724.3495977,Ouyang,redteaming,bai2022training}, the motivation for KL penalty is not widely understood.

In this paper, we discuss an underappreciated perspective on KL-regularised RL --- the objective employed by RL from human feedback for fine-tuning LMs --- which explains its empirical success. We start with describing a problem that arises from naively applying the standard RL objective: distribution collapse. The optimal policy under the RL objective would be a minimal-entropy LM generating a small set of sequences that obtain the highest reward. Then, we discuss how KL-regularised RL avoids distribution collapse due to its KL penalty. This constraint, we argue, transforms the problem from RL to Bayesian inference: updating a prior to conform with evidence provided by the reward. The Bayesian perspective moves KL-regularised RL closer to other divergence-minimisation-based approaches to fine-tuning LMs \citep{khalifa_2021} and, more broadly, to other divergence-minimisation-based accounts of control \citep{cai,hafner2020action}. These divergence minimisation approaches naturally avoid the distribution collapse problem because they formalize the agent as a generative model. In contrast, RL avoids distribution collapse only with reward functions that make it equivalent to divergence minimisation. Therefore, we conclude, RL is not an adequate formal framework for problems such as fine-tuning LMs.

\section{Fine-tuning language models using standard RL and distribution collapse}

Let $\mathcal{X}$ be the set of sequences of tokens from some vocabulary. An LM $\pi$ can be seen as a probability distribution over $\mathcal{X}$. While most modern LMs are autoregressive, for simplicity we will only talk about full sequences, e.g. $\pi(x)$ denotes the probability of a sequence $x\in\mathcal{X}$. Similarly, a reward function $r$ assigns sequences $x\in\mathcal{X}$ with scalar rewards. In practice, $r(x)$ could represent human preferences we would like $\pi$ to be aligned with, e.g. a non-offensiveness reward would assign low values to sequences that are offensive.

If $\pi_\theta$ is our parametric LM (with parameters $\theta$), the RL objective for fine-tuning it with our reward function $r$ is just the reward expected under LM distribution:
\begin{equation}
\label{rl}
    J_\text{RL}(\theta) = \mathbb{E}_{x\sim\pi_\theta} r(x).
\end{equation}
Intuitively, maximising $J_\text{RL}(\theta)$ means sampling a number of sequences from the LM and rewarding the LM for good sequences and penalising for bad ones (e.g. offensive sentences).


The problem with the RL objective is that it treats the LM as a policy, not as a generative model. While a generative model is supposed to capture a diverse distribution of samples, a policy is supposed to chose the optimal action. Since we don’t have a notion of state for LMs, the RL objective reduces to searching for $x^*$, the sequence with highest reward. If there is one, the optimal policy $\pi^*$ is a degenerate, deterministic generative model that puts entire probability mass on that single sequence:
\begin{equation}
    \pi^* = \text{argmax}_\theta J_\text{RL}(\theta) = \delta_{x^*},
\end{equation}
where $\delta_{x^*}$ is a Dirac delta distribution centred on $x^*$. If there are multiple optimal sequences  $x^*$, probability mass would be put only on them.

This failure mode is not purely theoretical. Empirically, distribution collapse induced by maximising reward manifests as decreased fluency and diversity of samples from the LM, which can be measured in terms of perplexity, entropy and the frequency of repetitions. Degeneration of this kind was observed in multiple language generation tasks ranging from translation \citep{choshen2019weaknesses}, summarisation \citep{PaulusXS18}, story generation \citep{RL_TambwekarDMMHR19}, video captioning \citep{PasunuruB17}, dialogue \citep{KL_jaquesK19}, to code generation \citep{korbak2021energybased} and LM debiasing \citep{khalifa_2021}.

While the distribution collapse problem is exacerbated by RL failure modes such as insufficient exploration or reward hacking, it is distinct from exploration-exploitation trade-off or reward misspecification. Even with perfect exploration (if we sampled sequences uniformly from $\mathcal{X}$ as opposed to sampling from $\pi_\theta$), the optimal policy will still put all probability mass on $x^*$. Similarly, even if $r$ perfectly captures human preferences across the whole space of possible sequences $\mathcal{X}$ and if $x^*$ is truly the best thing, we still would not want the LM to generate only $x^*$.\footnote{There is a case to be made that in conditional generation (e.g. translation or summarisation) one really cares \emph{only} about the single best output for a given context (e.g. summary of a document). There are still, however, substantial benefits of caring about distributional aspects in conditional generation. First, when the LM produces a full distribution, we are able to measure its uncertainty. For larger models, these uncertainty estimates happen to be well-calibrated and allow for safer deployment in high-stakes scenarios \citep{mostly_know}. Second, MAP estimates of the output distribution (the single, most likely output) are frequently of poor quality and can be substantially improved upon with decoding procedures taking into account the entire distribution, e.g. minimum Bayes risk in translation \citep{eikema-aziz-2020-map} or self-consistency chain-of-thought in question answering \citep{self-consistency}.  \citet{cascades} provide a unifying perspective on multi-step generation as latent variable modelling.} Essentially, the distribution collapse problem arises from the fact that the RL objective for LM alignment is flawed: it does not care about preserving distributional properties of an LM and will always penalise the LM for putting any probability mass on non-optimal sequences until the LM collapses into a degenerate distribution.

\section{Fine-tuning language models via KL-regularised RL}

There is an obvious solution to the distribution collapse problem: including preserving distributional properties of an LM as part of  the reward function. The notion of preserving distributional properties of an LM $\pi_\theta$ can be formalised as penalising for Kullback-Leibler (KL) divergence between $\pi_\theta$ and some other, pretrained LM $\pi_0$. Typically, $\pi_\theta$ is initialised to $\pi_0$ and then fine-tuned to maximise the following objective:
\begin{equation}
\label{KL-RL1}
    J_\text{KL-RL}(\theta) = \mathbb{E}_{x\sim\pi_\theta} [r(x)] - \beta D_\text{KL}(\pi_\theta,\pi_0).
\end{equation}
The first term in the right-hand side of \eqref{KL-RL1} is equivalent to $J_\text{RL}(\theta)$ in \eqref{rl} while the second additionally constrains $\pi_\theta$ to stay close (in terms of KL) to $\pi_0$. Almost always some reward needs to be sacrificed for that; the coefficient $\beta$ determines the trade-off of how much reward is needed to justify departing from $\pi_0$ by a certain distance. This objective is commonly used as part of a popular recipe for fine-tuning LMs termed ``RL from Human Feedback'' (RLHF) and works surprisingly well in practice \citep{Ziegler19,10.5555/3495724.3495977,redteaming,bai2022training}. Earlier approaches to fine-tuning LMs employing this objective used the called it ``conservative fine-tuning'' \citep{KL_Jaques17} or KL-control \cite{KL_jaquesK19}. Here, we focus only on the policy optimisation part of this setup, which we term ``KL-regularised RL''.

The KL-regularised RL objective \eqref{KL-RL1} can easily be reformulated as just expected reward as in \eqref{rl}. We only have to define a new reward function $r'_\theta(x)$ which incorporates the original reward $r$ and the KL penalty, using the definition of KL divergence:
\begin{equation}
\begin{aligned}
\label{KL-RL2}
    J_\text{KL-RL}(\theta) &= \mathbb{E}_{x\sim\pi_\theta} [r'_\theta(x)], \text{where}  \\
r'_\theta(x) &= r(x) + \beta(\log \pi_0(x) - \log \pi_\theta(x)).
\end{aligned}
\end{equation}
But is framing the maximisation of \eqref{KL-RL2} as RL really necessary?  In the next section, we will develop an alternative view of this objective -- as an approximate solution to a Bayesian inference problem -- and argue that it is a more appealing framing.

\section{KL-regularised RL as variational inference}

Fine-tuning a pretrained LM $\pi_0$ to align with preferences encoded by a reward function $r$ is essentially a Bayesian inference problem. Intuitively, Bayesian inference is the problem of updating a distribution to conform with new evidence. In our setting, we’re updating $\pi_\theta$, which is initially equal to a prior $\pi_0$ to conform with evidence provided by the assumption that $\pi_\theta$ is optimal in terms of $r$. A reward function can be represented as a distribution over $\mathcal{X}$ that makes high-reward sequences more likely that low-reward sequences. A simple way of doing that is exponentiating reward $r$ and renormalizing it. Then, the posterior is given by:
\begin{equation}
    \pi^*_\text{KL-RL}(x) = \frac{1}{Z}\pi_0(x)\exp(r(x)/\beta),
\end{equation}
where $\pi_0$ is the prior, $\exp(r(x)/\beta)$ is the evidence provided by the reward function (scaled by temperature $\beta$) and $Z$ is a constant ensuring that $\pi^*_\text{KL-RL}$ is a normalised probability distribution. $\pi^*_\text{KL-RL}$ represents a version of $\pi_0$ updated to account for the reward $r$. As we demonstrate in the Appendix, it also happens to coincide with the optimal policy for $J_\text{KL-RL}$:

\begin{equation}
    \pi^*_\text{KL-RL} = \text{argmax}_\theta J_\text{KL-RL}(\theta)
\end{equation}

Moreover, the KL-regularised RL objective can be cast as minimising the KL divergence between the LM $\pi_\theta$ and this target distribution $\pi^*_\text{KL-RL}$:
\begin{equation}
    J_\text{KL-RL}(\theta) \propto -D_\text{KL}(\pi_\theta, \pi^*_\text{KL-RL})
\end{equation}

This divergence is different from the KL penalty term $D_\text{KL}(\pi_\theta,\pi_0)$ in \eqref{KL-RL1}. Minimising this new divergence coincides with a variational inference \citep{Blei_2017}, a well-known approach to approximating Bayesian inference. More formally, $J_\text{KL-RL}(\theta)$ is the evidence lower bound (ELBO) on the log likelihood of $\pi_\theta$ being optimal under $r$, assuming a prior $\pi_0$. Minimising this bound makes $\pi_\theta$ approximate the true posterior $\pi^*_\text{KL-RL}$. A derivation of these equalities can be found in the Appendix below.

Why is this picture insightful? It explains where the KL penalty term $\beta D_\text{KL}(\pi_\theta,\pi_0)$ in KL-regularised RL's original objective comes from. It is necessary to transform the problem from RL to minimising a divergence from a target distribution $\pi^*_\text{RLKL}$. This in turn makes the distributional character of an LM a first-class citizen which explains why KL-regularised RL is able to maintain the fluency and diversity of the original LM $\pi_0$.

\section{Separation of modelling and inference}

The Bayesian perspective suggests that that aligning an LM with task preferences is a two-step process. It consists of, first, defining a distribution specifying the desired behaviour of your LM, and second, solving the problem of sampling from that posterior. These two steps roughly correspond to modelling and inference in probabilistic programming \citep{dippl}. Modelling is encoding knowledge in probabilistic terms (usually by defining a probabilistic graphical model) while inference corresponds to using this model to answer queries. It is hard to overstate how useful — theoretically and practically — separating these two concerns could be. Let us discuss these two steps, separately, below.

\paragraph{Modelling} The LM is natively a probability distribution and autoregressive models allow for both sampling and evaluating likelihoods. Therefore, most modelling decisions are usually around interpreting task preferences in probabilistic terms. Turning a reward function $r$ into a distribution by exponentiating it ($\frac{1}{Z}\exp(r(x)$) is the standard approach, but there are others. In some cases, task preferences can be binary, for instance a dialogue system might be required to \emph{never} generate a curse word (but is free to behave normally otherwise). Then, following \cite{khalifa_2021}, one could define $\pi^*(x) = \frac{1}{Z}\pi_0(x)b(x)$, where $b(x) = 1$ if $x$ contains a curse and $0$ otherwise. Then, sequences $x$ containing curses have probability zero according to $\pi^*$ (hence $\pi^*$ is non-cursing) but all other strings keep the original probability $\pi_0(x)$ up to $Z$ (hence no degeneration).

\paragraph{Inference} The posteriors mentioned above are generally non-parametric: they might lie outside the class of probability distributions representable by parametric LMs. Designing an algorithm able to generate samples matching this posterior distribution constitute the inference problem. Broadly, there are two families of algorithms for inference on probabilistic graphical models: variational inference and sampling-based approaches. Variational inference tries to find the set of weights $\theta$ that give rise to a distribution $\pi_\theta$ closest (in terms of KL) to the true posterior. Sampling-based techniques, such as MCMC \citep{brooks2011handbook}, do not represent the true posterior explicitly, but compute samples from a distribution resembling the true posterior. In the previous section, we have shown that KL-regularised RL corresponds to inference via variational inference. But sampling-based inference algorithms also have analogues for LMs in decoding-time methods. Decoding-time methods boil down to simulating a posterior, aligned LM $\pi^*$ by modifying the generation procedure applied on top of the original LM $\pi_0$. The simplest example of that is filtering (also known as rejection sampling): if the LM generates an unacceptable sample, it is discarded and a new sample is generated \citep{recipes}. More elaborate decoding-time methods include weighted decoding \citep{see-etal-2019-makes} or PPLM \citep{pplm}.

To summarise, the Bayesian view provides a unifying perspective on fine-tuning and decoding-time approaches to LM alignment. They mirror variational inference and sampling-based inference algorithms for probabilistic graphical models. But a more fundamental advantage, to our mind, is the separation of concerns between defining a desired behaviour of an LM and approximating it. The choice of posterior is independent of how it is going to be approximated. This, in turn, separates two failure modes: misspecifying the model (i.e. not capturing task preferences) and failing to approximate the model well enough.

\section{Is RL a good framework for fine-tuning language models?}

There is a family of other divergence minimisation approaches to fine-tuning LMs which are not equivalent to RL. Take Generative Distributional Control (GDC) \citep{khalifa_2021,pmlr-v162-korbak22a}, an approach to fine-tuning LMs that obtains results comparable with KL-regularised RL but minimises a slightly different divergence (forward as opposed to reverse KL). However, $J_\text{GDC}(\theta)$ is no longer equivalent to RL \citep{korbak2022on} because the expectation in forward KL divergence is with respect to a $\pi^*_\text{KL-RL}$, not $\pi_\theta$. Similarly, standard supervised training objective can be seen as minimising $D_\text{KL}(\pi^*_\text{MLE}, \pi_\theta)$, a divergence from the empirical distribution $\pi^*_\text{MLE}$ provided by the training set.

One can therefore mount a double dissociation argument in favour of the divergence minimisation perspective on KL-regularised RL: RL without KL divergence minimisation leads to degeneration while KL divergence minimisation without RL works well. Therefore, it is the KL divergence minimisation aspect of KL-regularised RL that seems to account for its success, not the reward maximisation aspect. In consequence, calling it RL is just a redescription of it that happens to be correct under a particular choice of reward function $r'_\theta$. However, this redescription does not provide motivation for this choice of $r'_\theta$ and does not hold for alternative divergence minimisation approaches to fine-tuning LMs such as GDC \cite{khalifa_2021}.

The divergence minimisation perspective on KL-regularised RL we presented stems from a general framework known as control as inference \cite{cai}. Control as inference provides a formalisation of intelligent decision making as inference on a probabilistic graphical model representing the agent, its preferences and environmental dynamics. While control as inference is typically considered with graphical models parameterised to make it equivalent to RL, it does not have to be. Moreover, there are frameworks such as active inference \citep{friston2010action,BUCKLEY201755} or action and perception as divergence minimisation \citep{hafner2020action} that further generalise control as inference to a principle of minimising the KL divergence from a probability distribution representing desired behaviour of the agent. In contrast with RL, they conceptualise the agent as a generative model, not as a decision rule represented as a probability distribution out of convenience. Therefore, they naturally avoid the distribution collapse problem and preserve the distributional properties of the agent. What if RL simply isn’t an adequate formal framework for problems such as aligning LMs?

\section*{Limitations}

In the paper, we discussed some limitations of standard approaches to using RL for fine-tuning LMs and sketched an alternative framing -- based on Bayesian inference -- of RLHF, a commonly used approach to RL fine-tuning. However, our discussion itself is limited in scope as we do not cover other shortcomings of RLHF and our own Bayesian proposal is not devoid of weaknesses. We take advantage of this section to examine these two sets of limitations.

\paragraph{Other limitations of RLHF} RLHF consists of (i) training a reward model to predict which of two texts a human
prefers and (ii) fine-tuning a pretrained LM to maximise reward given by the reward model. Our discussion focused on (ii) and took (i) as a given. But a reward model is always a proxy for the underlying task preferences and is limited in its ability to fit human feedback. Reward models are vulnerable to adversarial examples
\citep{https://doi.org/10.48550/arxiv.1606.04435,https://doi.org/10.48550/arxiv.1702.08138} and LMs optimised against them can exploit these adversarial examples \citep{pan2022effects}.

Moreover, training a reward model involves a multitude of non-technical design choices that shape the reward function the LM is optimised against. These design decisions involve data curation, annotation guideline preparation as well as annotator selection and compensation. Unintended bias can be introduced at each of these stages. For instance, crowdsource workers might be biased towards particular language varieties \citep{sap-etal-2019-risk}. More generally, preferences elicited from crowsource workers might not represent the preferences of the general population due to selection effects. For instance, most studies using RLHF recruits crowdsource workers either solely from the United States \citep{bai2022training} or from United States and Southeastern Asia \citep{10.5555/3495724.3495977,Ouyang}. Crowdsource workers frequently disagree among themselves and with researchers conducting the study.\footnote{The annotator-annotator agreement rates are 68\% in \citep{10.5555/3495724.3495977} and 72\% in \citep{Ouyang} while the the annotator-researcher agreement rates are 77\% in \citep{10.5555/3495724.3495977} and 63\% in \citep{bai2022training}.} This diversity of preferences makes the notion of a ground truth for the reward model problematic; see \citep[sec. 5.3-5.3]{Ouyang} for an extended discussion and \citep{gabriel2020artificial} for a philosophical examination of the notion of ground truth human preferences.

\paragraph{Limitations of the Bayesian perspective} We argued that RL with KL penalties and, more broadly, aligning language models with human preferences, can be seen as Bayesian inference and that this perspective is a more insightful theoretical grounding for RLHF than the standard RL perspective. However, our proposal as laid down above is only preliminary and does not account for some empirical regularities found in RLHF experiments. For instance, \citet{bai2022training} found that expected reward $\mathbb{E}_{x\sim\pi_\theta} r(x)$ is approximately linear in $\sqrt{D_\text{KL}(\pi_\theta, \pi_0)}$ throughout RLHF training. The Bayesian perspective remains to be developed to explain why such a relationship holds. Moreover, the Bayesian perspective currently offers limited guidance for design choices in RLHF experiments such as hyperparameter selection.

\section*{Ethics statement}

Our paper is a contribution to important lines of work on social bias in large language models and on aligning artificial intelligence with human preferences. The first line of work is primarily concerned with risks associated with an over-representation of certain hegemonic (e.g. sexist, racist, homophobic) viewpoints and voices present in the training data for large language models, which consists primarily of crawled, uncurated user-generated content. Deploying language models exhibiting social biases poses a risk of amplifying and perpetuaing these biases \citep{ShengCNP_LM_bias19,blodgett-bias-survey,Bender_parrots}. The second line of work is concerned more broadly with ensuring that objectives that machine learning systems pursue are aligned with human values \citep{https://doi.org/10.48550/arxiv.1606.06565,russell2019human}. Large language models, due to their capabilities, can be a testbed for alignment techniques for future, more powerful machine learning systems \citep{lab,https://doi.org/10.48550/arxiv.2110.08300}. Research on RLHF for fine-tuning LMs -- such as our paper -- can therefore be motivated by both narrower (social bias) and broader (alignment) considerations. As a theoretical contribution, our paper is not expected to pose significant risk. However, RLHF is a dual use technology: it can be diverted to malicious uses such as spreading misinformation or generating harmful content.

\section*{Acknowledgements}

Tomasz Korbak was supported by the Leverhulme Doctoral Scholarship and OpenPhilantropy. This paper benefited from discussions with Hady Elsahar, Marc Dymetman, Germán Kruszewski, Jérémy Scheurer, Kyle McDonell and Laria Reynolds.

\bibliography{references,additional_refs}

\begin{thebibliography}{44}
\expandafter\ifx\csname natexlab\endcsname\relax\def\natexlab#1{#1}\fi

\bibitem[{Amodei et~al.(2016)Amodei, Olah, Steinhardt, Christiano, Schulman,
  and Mané}]{https://doi.org/10.48550/arxiv.1606.06565}
Dario Amodei, Chris Olah, Jacob Steinhardt, Paul Christiano, John Schulman, and
  Dan Mané. 2016.
\newblock \href {https://doi.org/10.48550/ARXIV.1606.06565} {Concrete problems
  in ai safety}.

\bibitem[{Askell et~al.(2021)Askell, Bai, Chen, Drain, Ganguli, Henighan,
  Jones, Joseph, Mann, DasSarma, Elhage, Hatfield-Dodds, Hernandez, Kernion,
  Ndousse, Olsson, Amodei, Brown, Clark, McCandlish, Olah, and Kaplan}]{lab}
Amanda Askell, Yuntao Bai, Anna Chen, Dawn Drain, Deep Ganguli, Tom Henighan,
  Andy Jones, Nicholas Joseph, Ben Mann, Nova DasSarma, Nelson Elhage, Zac
  Hatfield-Dodds, Danny Hernandez, Jackson Kernion, Kamal Ndousse, Catherine
  Olsson, Dario Amodei, Tom Brown, Jack Clark, Sam McCandlish, Chris Olah, and
  Jared Kaplan. 2021.
\newblock \href {https://doi.org/10.48550/ARXIV.2112.00861} {A general language
  assistant as a laboratory for alignment}.

\bibitem[{Bai et~al.(2022)Bai, Jones, Ndousse, Askell, Chen, DasSarma, Drain,
  Fort, Ganguli, Henighan, Joseph, Kadavath, Kernion, Conerly, El-Showk,
  Elhage, Hatfield-Dodds, Hernandez, Hume, Johnston, Kravec, Lovitt, Nanda,
  Olsson, Amodei, Brown, Clark, McCandlish, Olah, Mann, and
  Kaplan}]{bai2022training}
Yuntao Bai, Andy Jones, Kamal Ndousse, Amanda Askell, Anna Chen, Nova DasSarma,
  Dawn Drain, Stanislav Fort, Deep Ganguli, Tom Henighan, Nicholas Joseph,
  Saurav Kadavath, Jackson Kernion, Tom Conerly, Sheer El-Showk, Nelson Elhage,
  Zac Hatfield-Dodds, Danny Hernandez, Tristan Hume, Scott Johnston, Shauna
  Kravec, Liane Lovitt, Neel Nanda, Catherine Olsson, Dario Amodei, Tom Brown,
  Jack Clark, Sam McCandlish, Chris Olah, Ben Mann, and Jared Kaplan. 2022.
\newblock \href {http://arxiv.org/abs/2204.05862} {Training a helpful and
  harmless assistant with reinforcement learning from human feedback}.

\bibitem[{Bender et~al.(2021)Bender, Gebru, McMillan-Major, and
  Shmitchell}]{Bender_parrots}
Emily~M. Bender, Timnit Gebru, Angelina McMillan-Major, and Shmargaret
  Shmitchell. 2021.
\newblock \href {https://doi.org/10.1145/3442188.3445922} {On the dangers of
  stochastic parrots: Can language models be too big?}
\newblock In \emph{Proceedings of the 2021 ACM Conference on Fairness,
  Accountability, and Transparency}, FAccT '21, page 610–623, New York, NY,
  USA. Association for Computing Machinery.

\bibitem[{Blei et~al.(2017)Blei, Kucukelbir, and McAuliffe}]{Blei_2017}
David~M. Blei, Alp Kucukelbir, and Jon~D. McAuliffe. 2017.
\newblock \href {https://doi.org/10.1080/01621459.2017.1285773} {Variational
  inference: A review for statisticians}.
\newblock \emph{Journal of the American Statistical Association},
  112(518):859--877.

\bibitem[{Blodgett et~al.(2020)Blodgett, Barocas, Daum{\'e}~III, and
  Wallach}]{blodgett-bias-survey}
Su~Lin Blodgett, Solon Barocas, Hal Daum{\'e}~III, and Hanna Wallach. 2020.
\newblock \href {https://doi.org/10.18653/v1/2020.acl-main.485} {Language
  (technology) is power: A critical survey of {``}bias{''} in {NLP}}.
\newblock In \emph{Proceedings of the 58th Annual Meeting of the Association
  for Computational Linguistics}, pages 5454--5476, Online. Association for
  Computational Linguistics.

\bibitem[{Bowman(2021)}]{https://doi.org/10.48550/arxiv.2110.08300}
Samuel~R. Bowman. 2021.
\newblock \href {https://doi.org/10.48550/ARXIV.2110.08300} {The dangers of
  underclaiming: Reasons for caution when reporting how nlp systems fail}.

\bibitem[{Brooks et~al.(2011)Brooks, Gelman, Jones, and
  Meng}]{brooks2011handbook}
Steve Brooks, Andrew Gelman, Galin Jones, and Xiao-Li Meng. 2011.
\newblock \emph{Handbook of Markov Chain Monte Carlo}.
\newblock CRC press.

\bibitem[{Brown et~al.(2020)Brown, Mann, Ryder, Subbiah, Kaplan, Dhariwal,
  Neelakantan, Shyam, Sastry, Askell, Agarwal, Herbert{-}Voss, Krueger,
  Henighan, Child, Ramesh, Ziegler, Wu, Winter, Hesse, Chen, Sigler, Litwin,
  Gray, Chess, Clark, Berner, McCandlish, Radford, Sutskever, and
  Amodei}]{gpt3}
Tom~B. Brown, Benjamin Mann, Nick Ryder, Melanie Subbiah, Jared Kaplan,
  Prafulla Dhariwal, Arvind Neelakantan, Pranav Shyam, Girish Sastry, Amanda
  Askell, Sandhini Agarwal, Ariel Herbert{-}Voss, Gretchen Krueger, Tom
  Henighan, Rewon Child, Aditya Ramesh, Daniel~M. Ziegler, Jeffrey Wu, Clemens
  Winter, Christopher Hesse, Mark Chen, Eric Sigler, Mateusz Litwin, Scott
  Gray, Benjamin Chess, Jack Clark, Christopher Berner, Sam McCandlish, Alec
  Radford, Ilya Sutskever, and Dario Amodei. 2020.
\newblock \href {http://arxiv.org/abs/2005.14165} {Language models are few-shot
  learners}.
\newblock \emph{CoRR}, abs/2005.14165.

\bibitem[{Buckley et~al.(2017)Buckley, Kim, McGregor, and Seth}]{BUCKLEY201755}
Christopher~L. Buckley, Chang~Sub Kim, Simon McGregor, and Anil~K. Seth. 2017.
\newblock \href {https://doi.org/https://doi.org/10.1016/j.jmp.2017.09.004}
  {The free energy principle for action and perception: A mathematical review}.
\newblock \emph{Journal of Mathematical Psychology}, 81:55--79.

\bibitem[{Choshen et~al.(2019)Choshen, Fox, Aizenbud, and
  Abend}]{choshen2019weaknesses}
Leshem Choshen, Lior Fox, Zohar Aizenbud, and Omri Abend. 2019.
\newblock On the weaknesses of reinforcement learning for neural machine
  translation.
\newblock \emph{arXiv preprint arXiv:1907.01752}.

\bibitem[{Dathathri et~al.(2019)Dathathri, Madotto, Lan, Hung, Frank, Molino,
  Yosinski, and Liu}]{pplm}
Sumanth Dathathri, Andrea Madotto, Janice Lan, Jane Hung, Eric Frank, Piero
  Molino, Jason Yosinski, and Rosanne Liu. 2019.
\newblock \href {https://doi.org/10.48550/ARXIV.1912.02164} {Plug and play
  language models: A simple approach to controlled text generation}.

\bibitem[{Dohan et~al.(2022)Dohan, Xu, Lewkowycz, Austin, Bieber, Lopes, Wu,
  Michalewski, Saurous, Sohl-dickstein, Murphy, and Sutton}]{cascades}
David Dohan, Winnie Xu, Aitor Lewkowycz, Jacob Austin, David Bieber,
  Raphael~Gontijo Lopes, Yuhuai Wu, Henryk Michalewski, Rif~A. Saurous, Jascha
  Sohl-dickstein, Kevin Murphy, and Charles Sutton. 2022.
\newblock \href {https://doi.org/10.48550/ARXIV.2207.10342} {Language model
  cascades}.

\bibitem[{Eikema and Aziz(2020)}]{eikema-aziz-2020-map}
Bryan Eikema and Wilker Aziz. 2020.
\newblock \href {https://doi.org/10.18653/v1/2020.coling-main.398} {Is {MAP}
  decoding all you need? the inadequacy of the mode in neural machine
  translation}.
\newblock In \emph{Proceedings of the 28th International Conference on
  Computational Linguistics}, pages 4506--4520, Barcelona, Spain (Online).
  International Committee on Computational Linguistics.

\bibitem[{Friston et~al.(2010)Friston, Daunizeau, Kilner, and
  Kiebel}]{friston2010action}
Karl~J Friston, Jean Daunizeau, James Kilner, and Stefan~J Kiebel. 2010.
\newblock Action and behavior: a free-energy formulation.
\newblock \emph{Biological cybernetics}, 102(3):227--260.

\bibitem[{Gabriel(2020)}]{gabriel2020artificial}
Iason Gabriel. 2020.
\newblock Artificial intelligence, values, and alignment.
\newblock \emph{Minds and machines}, 30(3):411--437.

\bibitem[{Gehman et~al.(2020)Gehman, Gururangan, Sap, Choi, and Smith}]{rtp}
Samuel Gehman, Suchin Gururangan, Maarten Sap, Yejin Choi, and Noah~A. Smith.
  2020.
\newblock \href {https://doi.org/10.48550/ARXIV.2009.11462}
  {Realtoxicityprompts: Evaluating neural toxic degeneration in language
  models}.

\bibitem[{Goodman and Stuhlm\"{u}ller(2014)}]{dippl}
Noah~D Goodman and Andreas Stuhlm\"{u}ller. 2014.
\newblock {The Design and Implementation of Probabilistic Programming
  Languages}.
\newblock \url{http://dippl.org}.
\newblock Accessed: 2022-6-17.

\bibitem[{Grosse et~al.(2016)Grosse, Papernot, Manoharan, Backes, and
  McDaniel}]{https://doi.org/10.48550/arxiv.1606.04435}
Kathrin Grosse, Nicolas Papernot, Praveen Manoharan, Michael Backes, and
  Patrick McDaniel. 2016.
\newblock \href {https://doi.org/10.48550/ARXIV.1606.04435} {Adversarial
  perturbations against deep neural networks for malware classification}.

\bibitem[{Hafner et~al.(2020)Hafner, Ortega, Ba, Parr, Friston, and
  Heess}]{hafner2020action}
Danijar Hafner, Pedro~A. Ortega, Jimmy Ba, Thomas Parr, Karl Friston, and
  Nicolas Heess. 2020.
\newblock \href {http://arxiv.org/abs/2009.01791} {Action and perception as
  divergence minimization}.

\bibitem[{Hosseini et~al.(2017)Hosseini, Kannan, Zhang, and
  Poovendran}]{https://doi.org/10.48550/arxiv.1702.08138}
Hossein Hosseini, Sreeram Kannan, Baosen Zhang, and Radha Poovendran. 2017.
\newblock \href {https://doi.org/10.48550/ARXIV.1702.08138} {Deceiving google's
  perspective api built for detecting toxic comments}.

\bibitem[{Jaques et~al.(2019)Jaques, Ghandeharioun, Shen, Ferguson, Lapedriza,
  Jones, Gu, and Picard}]{KL_jaquesK19}
Natasha Jaques, Asma Ghandeharioun, Judy~Hanwen Shen, Craig Ferguson,
  {\`{A}}gata Lapedriza, Noah Jones, Shixiang Gu, and Rosalind~W. Picard. 2019.
\newblock \href {http://arxiv.org/abs/1907.00456} {Way off-policy batch deep
  reinforcement learning of implicit human preferences in dialog}.
\newblock \emph{CoRR}, abs/1907.00456.

\bibitem[{Jaques et~al.(2017)Jaques, Gu, Bahdanau, Hern{\'{a}}ndez{-}Lobato,
  Turner, and Eck}]{KL_Jaques17}
Natasha Jaques, Shixiang Gu, Dzmitry Bahdanau, Jos{\'{e}}~Miguel
  Hern{\'{a}}ndez{-}Lobato, Richard~E. Turner, and Douglas Eck. 2017.
\newblock \href {http://proceedings.mlr.press/v70/jaques17a.html} {Sequence
  tutor: Conservative fine-tuning of sequence generation models with
  kl-control}.
\newblock In \emph{Proceedings of the 34th International Conference on Machine
  Learning, {ICML} 2017, Sydney, NSW, Australia, 6-11 August 2017}, volume~70
  of \emph{Proceedings of Machine Learning Research}, pages 1645--1654. {PMLR}.

\bibitem[{Kadavath et~al.(2022)Kadavath, Conerly, Askell, Henighan, Drain,
  Perez, Schiefer, Dodds, DasSarma, Tran-Johnson, Johnston, El-Showk, Jones,
  Elhage, Hume, Chen, Bai, Bowman, Fort, Ganguli, Hernandez, Jacobson, Kernion,
  Kravec, Lovitt, Ndousse, Olsson, Ringer, Amodei, Brown, Clark, Joseph, Mann,
  McCandlish, Olah, and Kaplan}]{mostly_know}
Saurav Kadavath, Tom Conerly, Amanda Askell, Tom Henighan, Dawn Drain, Ethan
  Perez, Nicholas Schiefer, Zac~Hatfield Dodds, Nova DasSarma, Eli
  Tran-Johnson, Scott Johnston, Sheer El-Showk, Andy Jones, Nelson Elhage,
  Tristan Hume, Anna Chen, Yuntao Bai, Sam Bowman, Stanislav Fort, Deep
  Ganguli, Danny Hernandez, Josh Jacobson, Jackson Kernion, Shauna Kravec,
  Liane Lovitt, Kamal Ndousse, Catherine Olsson, Sam Ringer, Dario Amodei, Tom
  Brown, Jack Clark, Nicholas Joseph, Ben Mann, Sam McCandlish, Chris Olah, and
  Jared Kaplan. 2022.
\newblock \href {https://doi.org/10.48550/ARXIV.2207.05221} {Language models
  (mostly) know what they know}.

\bibitem[{Khalifa et~al.(2021)Khalifa, Elsahar, and Dymetman}]{khalifa_2021}
Muhammad Khalifa, Hady Elsahar, and Marc Dymetman. 2021.
\newblock \href {https://openreview.net/forum?id=jWkw45-9AbL} {A distributional
  approach to controlled text generation}.
\newblock In \emph{International Conference on Learning Representations}.

\bibitem[{Korbak et~al.(2021)Korbak, Elsahar, Dymetman, and
  Kruszewski}]{korbak2021energybased}
Tomasz Korbak, Hady Elsahar, Marc Dymetman, and Germán Kruszewski. 2021.
\newblock \href {http://arxiv.org/abs/2106.04985} {Energy-based models for code
  generation under compilability constraints}.

\bibitem[{Korbak et~al.(2022{\natexlab{a}})Korbak, Elsahar, Kruszewski, and
  Dymetman}]{pmlr-v162-korbak22a}
Tomasz Korbak, Hady Elsahar, German Kruszewski, and Marc Dymetman.
  2022{\natexlab{a}}.
\newblock \href {https://proceedings.mlr.press/v162/korbak22a.html}
  {Controlling conditional language models without catastrophic forgetting}.
\newblock In \emph{Proceedings of the 39th International Conference on Machine
  Learning}, volume 162 of \emph{Proceedings of Machine Learning Research},
  pages 11499--11528. PMLR.

\bibitem[{Korbak et~al.(2022{\natexlab{b}})Korbak, Elsahar, Kruszewski, and
  Dymetman}]{korbak2022on}
Tomasz Korbak, Hady Elsahar, Germán Kruszewski, and Marc Dymetman.
  2022{\natexlab{b}}.
\newblock \href {https://doi.org/10.48550/ARXIV.2206.00761} {On reinforcement
  learning and distribution matching for fine-tuning language models with no
  catastrophic forgetting}.

\bibitem[{Levine(2018)}]{cai}
Sergey Levine. 2018.
\newblock \href {https://doi.org/10.48550/ARXIV.1805.00909} {Reinforcement
  learning and control as probabilistic inference: Tutorial and review}.

\bibitem[{Lin et~al.(2021)Lin, Hilton, and Evans}]{lin2021truthfulqa}
Stephanie Lin, Jacob Hilton, and Owain Evans. 2021.
\newblock Truthfulqa: Measuring how models mimic human falsehoods.
\newblock \emph{arXiv preprint arXiv:2109.07958}.

\bibitem[{Ouyang et~al.(2022)Ouyang, Wu, Jiang, Almeida, Wainwright, Mishkin,
  Zhang, Agarwal, Slama, Ray, Schulman, Hilton, Kelton, Miller, Simens, Askell,
  Welinder, Christiano, Leike, and Lowe}]{Ouyang}
Long Ouyang, Jeff Wu, Xu~Jiang, Diogo Almeida, Carroll~L. Wainwright, Pamela
  Mishkin, Chong Zhang, Sandhini Agarwal, Katarina Slama, Alex Ray, John
  Schulman, Jacob Hilton, Fraser Kelton, Luke Miller, Maddie Simens, Amanda
  Askell, Peter Welinder, Paul Christiano, Jan Leike, and Ryan Lowe. 2022.
\newblock \href {https://doi.org/10.48550/ARXIV.2203.02155} {Training language
  models to follow instructions with human feedback}.

\bibitem[{Pan et~al.(2022)Pan, Bhatia, and Steinhardt}]{pan2022effects}
Alexander Pan, Kush Bhatia, and Jacob Steinhardt. 2022.
\newblock \href {http://arxiv.org/abs/2201.03544} {The effects of reward
  misspecification: Mapping and mitigating misaligned models}.

\bibitem[{Pasunuru and Bansal(2017)}]{PasunuruB17}
Ramakanth Pasunuru and Mohit Bansal. 2017.
\newblock \href {https://doi.org/10.18653/v1/d17-1103} {Reinforced video
  captioning with entailment rewards}.
\newblock In \emph{Proceedings of the 2017 Conference on Empirical Methods in
  Natural Language Processing, {EMNLP} 2017, Copenhagen, Denmark, September
  9-11, 2017}, pages 979--985. Association for Computational Linguistics.

\bibitem[{Paulus et~al.(2018)Paulus, Xiong, and Socher}]{PaulusXS18}
Romain Paulus, Caiming Xiong, and Richard Socher. 2018.
\newblock \href {https://openreview.net/forum?id=HkAClQgA-} {A deep reinforced
  model for abstractive summarization}.
\newblock In \emph{6th International Conference on Learning Representations,
  {ICLR} 2018, Vancouver, BC, Canada, April 30 - May 3, 2018, Conference Track
  Proceedings}. OpenReview.net.

\bibitem[{Perez et~al.(2022)Perez, Huang, Song, Cai, Ring, Aslanides, Glaese,
  McAleese, and Irving}]{redteaming}
Ethan Perez, Saffron Huang, Francis Song, Trevor Cai, Roman Ring, John
  Aslanides, Amelia Glaese, Nat McAleese, and Geoffrey Irving. 2022.
\newblock \href {https://doi.org/10.48550/ARXIV.2202.03286} {Red teaming
  language models with language models}.

\bibitem[{Russell(2019)}]{russell2019human}
Stuart Russell. 2019.
\newblock \href {https://books.google.com/books?id=M1eFDwAAQBAJ} {\emph{Human
  Compatible: Artificial Intelligence and the Problem of Control}}.
\newblock Penguin Publishing Group.

\bibitem[{Sap et~al.(2019)Sap, Card, Gabriel, Choi, and
  Smith}]{sap-etal-2019-risk}
Maarten Sap, Dallas Card, Saadia Gabriel, Yejin Choi, and Noah~A. Smith. 2019.
\newblock \href {https://doi.org/10.18653/v1/P19-1163} {The risk of racial bias
  in hate speech detection}.
\newblock In \emph{Proceedings of the 57th Annual Meeting of the Association
  for Computational Linguistics}, pages 1668--1678, Florence, Italy.
  Association for Computational Linguistics.

\bibitem[{See et~al.(2019)See, Roller, Kiela, and Weston}]{see-etal-2019-makes}
Abigail See, Stephen Roller, Douwe Kiela, and Jason Weston. 2019.
\newblock \href {https://doi.org/10.18653/v1/N19-1170} {What makes a good
  conversation? how controllable attributes affect human judgments}.
\newblock In \emph{Proceedings of the 2019 Conference of the North {A}merican
  Chapter of the Association for Computational Linguistics: Human Language
  Technologies, Volume 1 (Long and Short Papers)}, pages 1702--1723,
  Minneapolis, Minnesota. Association for Computational Linguistics.

\bibitem[{Sheng et~al.(2019)Sheng, Chang, Natarajan, and
  Peng}]{ShengCNP_LM_bias19}
Emily Sheng, Kai{-}Wei Chang, Premkumar Natarajan, and Nanyun Peng. 2019.
\newblock \href {https://doi.org/10.18653/v1/D19-1339} {The woman worked as a
  babysitter: On biases in language generation}.
\newblock In \emph{Proceedings of the 2019 Conference on Empirical Methods in
  Natural Language Processing and the 9th International Joint Conference on
  Natural Language Processing, {EMNLP-IJCNLP} 2019, Hong Kong, China, November
  3-7, 2019}, pages 3405--3410. Association for Computational Linguistics.

\bibitem[{Stiennon et~al.(2020)Stiennon, Ouyang, Wu, Ziegler, Lowe, Voss,
  Radford, Amodei, and Christiano}]{10.5555/3495724.3495977}
Nisan Stiennon, Long Ouyang, Jeff Wu, Daniel~M. Ziegler, Ryan Lowe, Chelsea
  Voss, Alec Radford, Dario Amodei, and Paul Christiano. 2020.
\newblock Learning to summarize from human feedback.
\newblock In \emph{Proceedings of the 34th International Conference on Neural
  Information Processing Systems}, NIPS'20, Red Hook, NY, USA. Curran
  Associates Inc.

\bibitem[{Tambwekar et~al.(2019)Tambwekar, Dhuliawala, Martin, Mehta, Harrison,
  and Riedl}]{RL_TambwekarDMMHR19}
Pradyumna Tambwekar, Murtaza Dhuliawala, Lara~J. Martin, Animesh Mehta, Brent
  Harrison, and Mark~O. Riedl. 2019.
\newblock \href {https://doi.org/10.24963/ijcai.2019/829} {Controllable neural
  story plot generation via reward shaping}.
\newblock In \emph{Proceedings of the Twenty-Eighth International Joint
  Conference on Artificial Intelligence, {IJCAI} 2019, Macao, China, August
  10-16, 2019}, pages 5982--5988. ijcai.org.

\bibitem[{Wang et~al.(2022)Wang, Wei, Schuurmans, Le, Chi, Narang, Chowdhery,
  and Zhou}]{self-consistency}
Xuezhi Wang, Jason Wei, Dale Schuurmans, Quoc Le, Ed~Chi, Sharan Narang,
  Aakanksha Chowdhery, and Denny Zhou. 2022.
\newblock \href {https://doi.org/10.48550/ARXIV.2203.11171} {Self-consistency
  improves chain of thought reasoning in language models}.

\bibitem[{Xu et~al.(2020)Xu, Ju, Li, Boureau, Weston, and Dinan}]{recipes}
Jing Xu, Da~Ju, Margaret Li, Y-Lan Boureau, Jason Weston, and Emily Dinan.
  2020.
\newblock \href {https://doi.org/10.48550/ARXIV.2010.07079} {Recipes for safety
  in open-domain chatbots}.

\bibitem[{Ziegler et~al.(2019)Ziegler, Stiennon, Wu, Brown, Radford, Amodei,
  Christiano, and Irving}]{Ziegler19}
Daniel~M. Ziegler, Nisan Stiennon, Jeffrey Wu, Tom~B. Brown, Alec Radford,
  Dario Amodei, Paul Christiano, and Geoffrey Irving. 2019.
\newblock \href {http://arxiv.org/abs/1909.08593} {Fine-tuning language models
  from human preferences}.
\newblock \emph{CoRR}, abs/1909.08593.

\end{thebibliography}
\bibliographystyle{acl_natbib}
\clearpage
\appendix
\section*{Appendix}
\label{sec:appendix}

Let’s assume we have a prior distribution over sequences of tokens $\pi_0(x)$ and a reward function $r$ which is (for technical reasons) always negative (from $-\infty$ to 0). We can also represent $r$ as a binary random variable $\mathcal{O}$ (the optimality variable). $\mathcal{O} = 1$ if a certain LM $\pi$ is optimal. We can define $\mathcal{O}$ in terms of $r$ as
\begin{equation}
    p(\mathcal{O}=1|x) = \exp(r(x)),
\end{equation}
which is normalised because $r(x)$ is always negative. For instance, if $r(x)$ is a log probability that a sequence $x$ is non-offensive, $p(\mathcal{O}=1|x)$ is a probability that $x$ is non-offensive and the marginal $p(\mathcal{O}=1)$ is the average offensiveness score of $\pi$ (or a probability that a random sample from $\pi$ is non-offensive). The problem of aligning LMs can be seen as inferring $p(x|\mathcal{O}=1)$, a distribution over sequences of tokens conditioned on being non-offensive. This can be computed by applying Bayes rule as

\begin{align}
p(x|\mathcal{O}=1) =& \frac{p(\mathcal{O}=1|x)p(x)}{p(\mathcal{O}=1)} \\ 
=& \frac{1}{Z}a(x)\exp(r(x)/\beta),
\end{align}
where we chose the prior $p(x)=\pi_0(x)$, redefined the marginal $p(\mathcal{O}=1)$ as the normalising constant $Z$, used the definition of $p(\mathcal{O}=1|x)$ and chose $\beta=1$.  $p(x|\mathcal{O}=1)$ here is equivalent to $\pi^*_\text{KL-RL}$, the optimal policy under  objective in \eqref{KL-RL1} (up to the choice of $\beta$ which can be absorbed into $r$).

 $p(x|\mathcal{O}=1)$ is a non-parametric distribution: it doesn’t have to lie in the family of distributions representable by a parametric model. In general, we’d like to find a parametric model $\pi_\theta$ closest to  $\pi^*_\text{KL-RL}$. This can be formalised as finding $\pi_\theta$ minimising $D_\text{KL}(\pi_\theta, \pi^*_\text{KL-RL})$. Here, however, we will derive this objective from a yet more general perspective: inferring a random latent variable $x$ that best explains the assumption that certain LM $\pi$ is optimal given a prior $\pi_0(x)$. This can be seen as maximising the log-likelihood of $\mathcal{O}=1$ via variational inference:

\begin{align}
    \log p(\mathcal{O}=1) &= \log \sum_x p(\mathcal{O}=1,x) \label{eq1} \\
&= \log \Big[ \sum_x p(\mathcal{O}=1|x)\pi_0(x) \Big]  \label{eq15}\\
&=\log \Big[\sum_x \pi_\theta(x) p(\mathcal{O}=1|x)\frac{\pi_0(x)}{\pi_\theta(x) } \Big] \label{eq2} \\
& \geq \sum_x \pi_\theta(x) \log \Big[ p(\mathcal{O}=1|x) \frac{\pi_0(x)}{\pi_\theta(x) }\Big] \label{eq3}\\
&=\mathbb{E}_{x\sim\pi_\theta} \log \Big[ \exp(r(x)) \frac{\pi_0(x)}{\pi_\theta(x) }\Big] \label{eq4}
\end{align}

In this derivation, we first introduce a latent variable $x$ using the sum rule of probability \eqref{eq1}, factorize a joint distribution \eqref{eq15}, introduce a variational distribution $\pi_\theta$ over that latent variable \eqref{eq2}, use Jensen’s inequality to obtain a bound (ELBo) \eqref{eq3} and, finally in  \eqref{eq4}, use the definition of $p(\mathcal{O}=1|x)$. This new bound can be alternatively expressed in two different ways:

\begin{equation}
        \mathbb{E}_{x\sim \pi_\theta} [r(x)] - D_\text{KL}(\pi_\theta,a), \label{KL-RL2a}
\end{equation}
\begin{equation}
        -\mathbb{E}_{x\sim \pi_\theta} \log\frac{\pi_\theta(x)}{\pi_0(x)\exp(r(x))}.\label{KL-RL3}
\end{equation}
\eqref{KL-RL2a} is just KL-regularised RL objective $J_\text{KL-RL}(\theta)$ with $\beta=1$. \eqref{KL-RL3} is proportional (up to a constant $-\log Z$) to negative $D_\text{KL}(\pi_\theta, \pi^*_\text{KL-RL})$, where $\pi^*_\text{KL-RL}=\frac{1}{Z}\pi_0(x)\exp(r(x))$ is the target distribution (or optimal policy for $J_\text{KL-RL}(\theta)$). Their equivalence proves that KL-regularised reward maximisation is equivalent to minimising divergence from $\pi^*_\text{KL-RL}$.

More broadly, the derivation above shows that $J_\text{KL-RL}(\theta)$ can be derived from first principles under a framework called control-as-inference \citep{cai}. The central idea here is to start from a Bayesian inference problem: inferring a distribution over $x$ (an LM) that reconciles the assumption that this LM is optimal ($p(\mathcal{O}=1)=1$, which plays we role of evidence) with a prior $\pi_0(x)$. KL-regularised RL arises as we solve this inference problem approximately via variational inference, i.e. by introducing a variational distribution $\pi_\theta$ and optimising it to maximise a lower bound on $\log p(\mathcal{O}=1)$.

\end{document}